\PassOptionsToPackage{hyphens}{url}
\RequirePackage[svgnames,table]{xcolor}
\documentclass[11pt,letterpaper,logo]{yalearxiv}

\usepackage[numbers,sort&compress]{natbib}
\usepackage{multirow,array,makecell,mathtools}
\usepackage{flafter,float}
\definecolor{RegisteredResult}{HTML}{C2185B}
\definecolor{MethodBlue}{HTML}{244A7C}
\definecolor{MethodTeal}{HTML}{138A83}
\definecolor{MethodGold}{HTML}{B7791F}
\newcommand{\result}[1]{\textbf{#1}}
\newcommand{\resulttext}[1]{#1}
\newcommand{\resultpm}[2]{\result{#1$\pm$#2}}
\newcommand{\sg}{\operatorname{sg}}

\newcommand{\softmax}{\operatorname{softmax}}
\newcommand{\ind}{\mathbb{I}}

\newtheorem{proposition}{Proposition}

\newcommand{\norm}[1]{\lVert#1\rVert}
\newcolumntype{Y}{>{\centering\arraybackslash}X}
\newcolumntype{L}[1]{>{\raggedright\arraybackslash}p{#1}}
\title{Selective Posterior Margin Regularization for Forward-Corrected Classification}
\runningtitle{Selective Posterior Margin Regularization}
\author{Zexing Zhang, Jichao Li, Tianyang Lei, XiongYi Lu, and Yang Kewei\\
College of Systems Engineering, National University of Defense Technology\\
Changsha 410073, China\\
\texttt{zhangzexing@nudt.edu.cn}, \texttt{lijichao09@nudt.edu.cn},
\texttt{leitianyang20@163.com}\\
\texttt{luxiongyi@nudt.edu.cn}, \texttt{kayyang27@nudt.edu.cn}}
\keywords{noisy-label learning, forward correction, posterior regularization, robust classification}

\hypersetup{
  colorlinks=true,
  linkcolor=blue!50!black,
  citecolor=blue!50!black,
  urlcolor=blue!50!black,
  pdftitle={Selective Posterior Margin Regularization for Forward-Corrected Classification},
  pdfauthor={Zexing Zhang, Jichao Li, Tianyang Lei, XiongYi Lu, Yang Kewei},
  pdfsubject={Learning with class-conditional label noise},
  pdfkeywords={noisy-label learning, forward correction, posterior regularization, robust classification}
}

\begin{document}
\begin{abstract}
Learning with class-conditional label noise often relies on a transition model from latent clean classes to observed annotations. Forward correction embeds this transition in the likelihood, yet finite-sample networks may still memorize corrupted labels. The corrected likelihood also induces a reverse posterior over the clean classes that could explain each annotation. When its leading class differs from the annotation, the model and transition matrix provide evidence against that annotation, but the leading alternatives can remain nearly tied. We introduce Selective Posterior Margin Regularization (SPMR), which preserves the Forward objective and converts this disagreement into a graded update on the clean classifier. SPMR selects the leading reverse-posterior class, scales a detached pairwise margin by the separation between the two leading posterior classes, and assigns correspondingly little influence to diffuse conflicts. The gap factorizes into transition-adjusted pairwise separation and the posterior mass carried by the leading pair. The active margin follows the locally minimum-norm logit direction that enlarges the selected pairwise margin. \resulttext{Across five known-transition benchmarks, SPMR improves full-length Forward by 2.5--7.0 percentage points and remains 0.7--2.5 percentage points above Forward with Mixup and early stopping. Matched interventions support distinct gains from the posterior-space coefficient, transition-adjusted target, and pairwise action. The same design transfers to estimated transitions, human annotations, architectural changes, and stronger Forward recipes.} The formulation uses latent-class evidence already available inside Forward correction without promoting every posterior conflict to a corrected label.
\end{abstract}

\maketitle

\section{Introduction}
Large visual datasets increasingly rely on crowdsourcing, web retrieval, or automated annotation. Their errors are rarely uniform across classes. A dog image is more likely to be labeled as a cat than as an airplane, for example. Class-conditional noise models capture this structure through a transition matrix that maps latent clean classes to observed annotations. Once this matrix is known or estimated, the corruption process can be incorporated directly into the training loss \cite{natarajan2013learning,patrini2017making}. The same concern is increasingly relevant to embodied learning, where policies combine heterogeneous robot trajectories with Internet-scale vision--language data \cite{zitkovich2023rt2,openx2024}.

Forward correction is among the most widely used realizations of this idea. It maps a classifier's clean-class probabilities through the transition matrix and maximizes the likelihood of the observed label. The corresponding population objective is statistically motivated, but its finite-sample behavior can be less favorable. Modern networks may continue fitting corrupted annotations after the useful structure has been learned, and recent results show that this late deterioration can occur even with an exact transition matrix \cite{feng2026deconstructing}. Strengthening the surrounding recipe with representation learning, augmentation, or early stopping helps, but does not answer what can be learned from the corrected likelihood itself.

\begin{figure}[t]
\centering
\includegraphics[width=\textwidth]{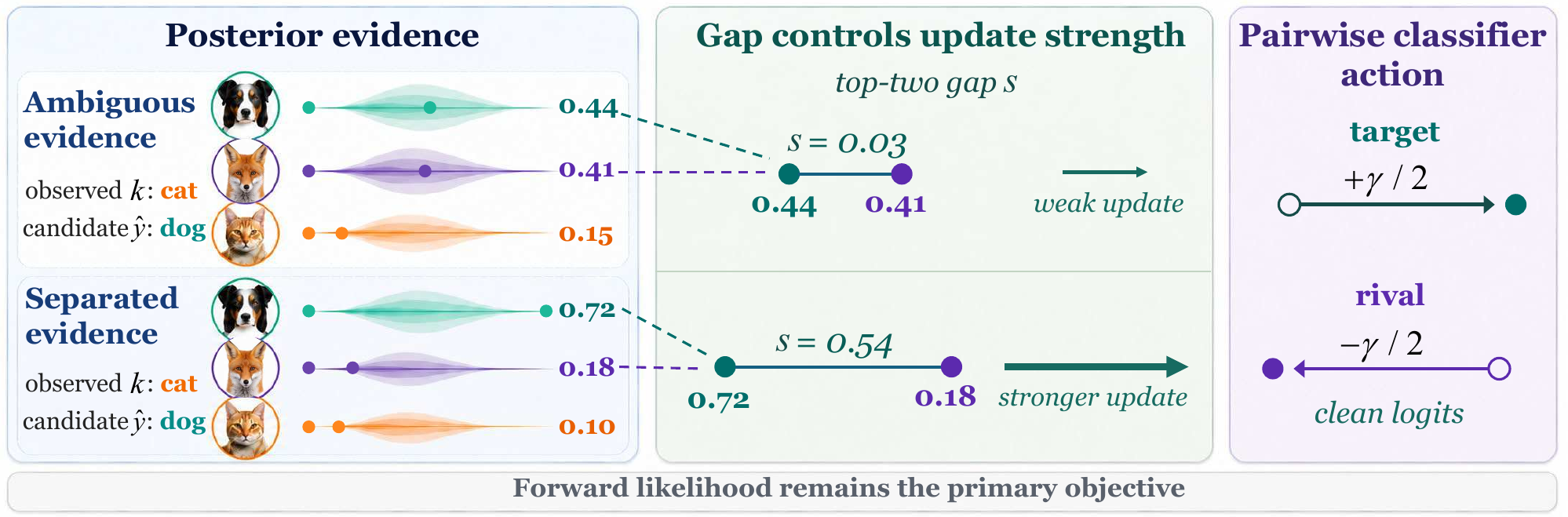}
\caption{\textbf{Posterior disagreement requires a second decision.} Both examples conflict with the observed annotation. The displayed posterior values illustrate how a small separation yields a weak intervention, whereas a clearly separated alternative receives a stronger clean-logit margin. Distances and distribution shapes are schematic; the annotated posterior values and gaps are exact.}
\label{fig:teaser}
\end{figure}

Forward computes more than an observed-label loss. For every training example it also induces a reverse posterior over the latent clean classes that could have generated the annotation. When the most probable clean class differs from the annotation, the model and transition matrix jointly favor an alternative explanation. This disagreement is appealing as a training signal, yet it is not automatically a corrected label. As illustrated in Figure~\ref{fig:teaser}, the leading alternatives may remain almost tied. Treating the winner as certain can convert a useful diagnostic into confirmation bias.

Posterior-based supervision must resolve three distinct questions. The posterior first supplies \emph{evidence} about whether the annotation remains plausible and whether one alternative is sufficiently separated. It then identifies a candidate \emph{target}, while the training objective determines the \emph{action} applied to the clean classifier. These roles are often changed together by pseudo-labeling or sample reweighting. Separating them makes it possible to determine whether a gain comes from the posterior evidence, the selected class, or the geometry of the additional update.

SPMR is a minimal realization of this decomposition. The Forward likelihood remains the primary objective. On posterior-conflict examples, SPMR uses the gap between the two leading reverse-posterior classes as a continuous coefficient and applies a detached pairwise margin to the clean logits used at test time. Ambiguous conflicts therefore receive little emphasis, whereas a clearly separated alternative receives a stronger update. This cross-space design lets the transition model decide how compelling the latent-class explanation is, while the auxiliary loss acts directly on the classifier whose predictions are ultimately evaluated.

Two structural observations make the design transparent. The posterior top-two gap separates into a transition-adjusted pairwise score and the posterior mass assigned to the leading pair. The pairwise hinge has an equally direct geometric interpretation. Among local logit changes that achieve a prescribed increase of the selected margin, its descent direction has minimum Euclidean norm. These properties do not claim global optimality. They identify what information the coefficient contains and what first-order action the auxiliary loss performs.

This decomposition also determines the empirical tests. We compare posterior separation with prediction-space and simpler confidence scores on the same conflict set and target. Under asymmetric noise, we replace only the target to identify the transition's contribution to class choice. Finally, practical auxiliary losses are complemented by detached linear surrogates that match each example's additional logit-gradient norm, leaving the first-order direction as the controlled difference. \resulttext{Relative to their strongest matched alternatives, posterior evidence, the pairwise direction, and transition-adjusted target improve accuracy by 1.34, 1.22, and 1.33 percentage points, respectively.} Transfer studies then examine estimated transitions, stronger Forward recipes, human annotations, and architectural changes.

\enlargethispage{1.2\baselineskip}
The main contributions are as follows.
\begin{itemize}
    \item We formulate auxiliary supervision inside Forward correction as an evidence--target--action problem and introduce SPMR, which preserves the corrected likelihood while adding a selective update to the clean classifier.
    \item We characterize posterior separation, the local geometry of the pairwise action, and the stability of transition-adjusted class comparisons under entrywise transition error.
    \item We evaluate each role with matched interventions, including equal-norm logit-gradient injections, and test whether the resulting gains persist across known and estimated transitions, stronger training recipes, human annotations, and architectural transfer.
\end{itemize}

\section{Related Work}
\paragraph{Loss correction and transition estimation.}
Noisy-label learning spans loss correction, robust objectives, sample selection, label revision, and semi-supervised training \cite{song2023survey}. Transition-based correction models how clean labels become observed annotations and modifies the likelihood accordingly \cite{natarajan2013learning,patrini2017making}. Neural variants have learned explicit noise layers, bootstrapped targets, or transition matrices from trusted and noisy data \cite{sukhbaatar2015training,goldberger2017noise,reed2015bootstrapping,hendrycks2018trusted,xia2019revision,yao2020dualt}. Noise-Bounded Forward changes the corrected loss near the probability boundary \cite{toner2023forward}, whereas FEC and JEC strengthen Forward through representation learning and Mixup \cite{feng2026deconstructing}. SPMR leaves the corrected likelihood unchanged and instead asks whether its reverse posterior supports a separate, local action on the clean classifier.

\paragraph{Sample selection and posterior targets.}
Many methods exploit the tendency of networks to fit regular structure before memorizing corrupted annotations. Co-teaching variants and JoCoR exchange or jointly select small-loss examples \cite{han2018coteaching,yu2019coteachingplus,wei2020jocor}. SELF, DivideMix, UniCon, and CrossSplit combine selection with self-ensembling, semi-supervised learning, contrastive representation learning, or independent training splits \cite{nguyen2020self,li2020dividemix,karim2022unicon,kim2023crosssplit}. Other approaches revise labels probabilistically or identify suspicious examples from margin and confidence dynamics \cite{arazo2019pencil,pleiss2020aum,northcutt2021confident}. A transition-corrected clean-class posterior has also been explored for sample selection and relabeling \cite{yao2022posteriorselection}, while CDRO constructs a robust target distribution around an estimated true-label posterior \cite{guo2024cdro}. SPMR neither partitions the dataset nor replaces the annotation. It uses relative posterior evidence to scale a pairwise constraint, and the same-active-set experiments separate that evidence from generic confidence selection.

\paragraph{Robust objectives and update control.}
Robust losses alter boundedness, symmetry, normalization, or saturation to limit corrupted-label influence \cite{gosh2017robust,zhang2018gce,wang2019sce,ma2020normalized,englesson2021gjs}. ELR and robust early learning use temporal behavior to resist late memorization \cite{liu2020elr,xia2021robustearly}, while augmentation and contrastive learning can improve the representation on which robust training operates \cite{zhang2018mixup,nishi2021augmentation,xue2022contrastive}. HMW derives a sample weight from hyperspherical feature margins, and OGC adapts a gradient-clipping threshold over training \cite{zhang2024hmw,ye2025ogc}. SPMR retains the Forward gradient and adds a target-specific pairwise direction. The norm-matched intervention therefore tests a distinction that loss replacement, sample weighting, and clipping do not isolate.

\paragraph{Positioning of the empirical comparison.}
Two-network and semi-supervised systems such as DivideMix, UniCon, and CrossSplit change the data partition, model count, and training objective, so their published results are not causal controls for an intervention inside Forward. Our common-protocol comparisons focus on HMW and OGC as neighboring weighting and gradient-control mechanisms, while the FEC/JEC experiments test whether SPMR remains complementary after the surrounding Forward recipe is strengthened.

\section{Method}
\begin{figure}[t]
\centering
\includegraphics[width=\textwidth]{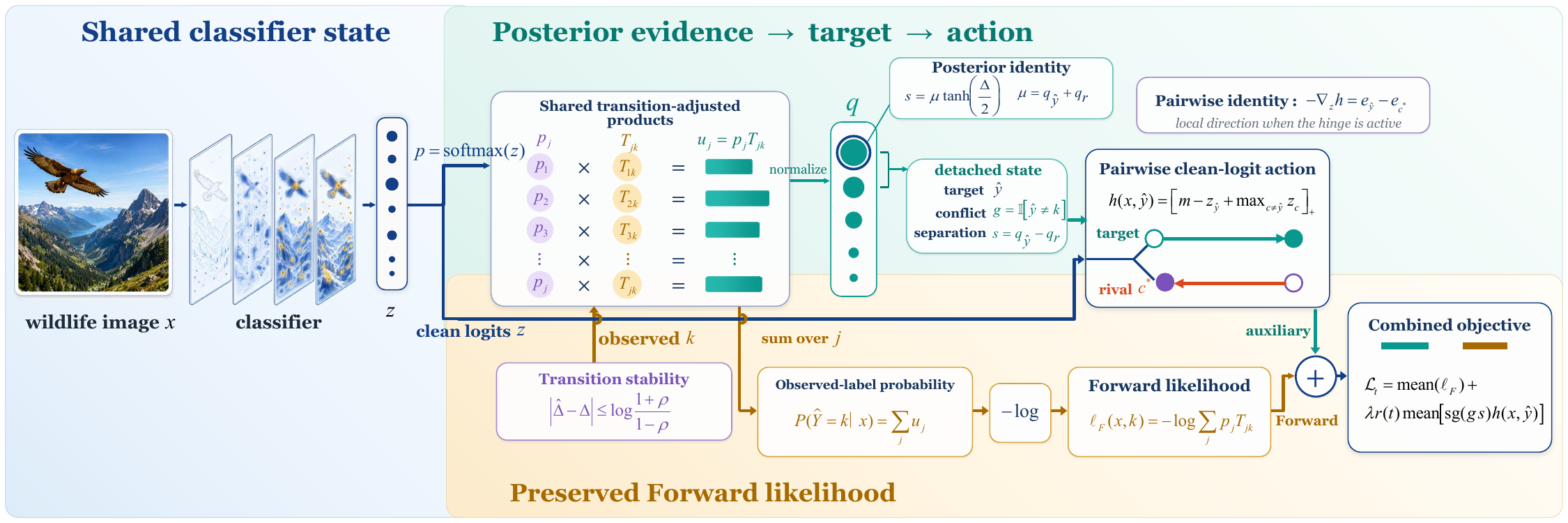}
\caption{\textbf{SPMR separates posterior evidence, target choice, and classifier action while preserving the Forward likelihood.} The compact lower row gives the two structural identities used in the analysis. Transition stability is developed in Eq.~\eqref{eq:perturb_bound}.}
\label{fig:overview}
\end{figure}

Figure~\ref{fig:overview} summarizes SPMR as a two-branch extension of Forward. The likelihood branch continues to explain the observed annotation through the transition matrix. The auxiliary branch reuses the resulting reverse posterior to quantify evidence, identify a candidate clean class, and apply a detached pairwise constraint to the test-time logits. This separation is central to the method. Posterior inference decides how compelling an alternative explanation is, whereas the auxiliary objective determines how the clean classifier should respond.

\subsection{Reverse-Posterior Evidence}
Let $Y\in[K]$ denote the latent clean label and $\widetilde Y\in[K]$ the observed label. Class-conditional noise is represented by $T_{jk}=P(\widetilde Y=k\mid Y=j)$, where clean classes index rows. A classifier produces logits $z=f_\theta(x)$ and clean probabilities $p=\softmax(z)$. For an example observed as class $k$, Forward correction and the corresponding reverse posterior are
\begin{equation}
\ell_F(x,k)=-\log\sum_j p_jT_{jk},\qquad
q_j=\frac{p_jT_{jk}}{\sum_c p_cT_{ck}}.
\label{eq:forward_reverse}
\end{equation}
The logit gradient is $\nabla_z\ell_F=p-q$. Since $q$ changes with the network, this identity describes the current training step rather than optimization against a fixed soft target.

Let $\hat y=\arg\max_j q_j$ and let $r$ index the second-largest posterior entry. We define
\begin{equation}
g=\ind[\hat y\neq k],\qquad s=q_{\hat y}-q_r.
\label{eq:evidence}
\end{equation}
The indicator identifies a posterior conflict, while $s$ measures how clearly the leading clean-class explanation separates from its closest alternative. The coefficient is continuous. A conflict with nearly tied alternatives produces a small update rather than a second hard selection threshold.

The gap admits a useful factorization. Define $a_j=z_j+\log T_{jk}$, $\Delta=a_{\hat y}-a_r$, and $\mu=q_{\hat y}+q_r$. Then
\begin{equation}
s=\mu\tanh(\Delta/2),\qquad
\Delta=(z_{\hat y}-z_r)+\log\frac{T_{\hat y k}}{T_{rk}}.
\label{eq:gap_factorization}
\end{equation}
The first factor records how much posterior mass is concentrated on the leading pair. The second is a bounded transformation of their transition-adjusted score difference. A large gap therefore requires both a decisive pairwise comparison and little probability mass left for other classes. The factorization follows directly from the softmax ratio, and the experiments below evaluate the factors and common confidence alternatives separately.

\subsection{Target and Pairwise Action}
The leading reverse-posterior class $\hat y$ supplies the candidate target. SPMR acts on the clean classifier through
\begin{equation}
h(x,\hat y)=\left[m-z_{\hat y}+\max_{c\neq\hat y}z_c\right]_+.
\label{eq:hinge}
\end{equation}
The margin is active only while $\hat y$ is less than $m$ above its strongest competitor. It therefore focuses on the comparison that currently prevents the selected class from being decisive.

\begin{proposition}
Let $c^*=\arg\max_{c\neq\hat y}z_c$. Among logit displacements that increase the current $\hat y$-versus-$c^*$ margin by at least $\gamma>0$, the minimum-$\ell_2$ displacement is $\frac{\gamma}{2}(e_{\hat y}-e_{c^*})$. The descent direction of an active hinge is parallel to this displacement.
\end{proposition}
The proposition gives a local interpretation of the action. Only the selected class and its current strongest competitor need to move, and they move symmetrically. This differs from hard-target cross-entropy, which redistributes probability across all classes. Indeed, minimizing $\norm{v}_2$ under $v_{\hat y}-v_{c^*}\geq\gamma$ gives $v_{\hat y}=\gamma/2$, $v_{c^*}=-\gamma/2$, and zero displacement in every other coordinate.

The complete training objective is
\begin{equation}
\mathcal L_t=\operatorname{mean}(\ell_F)+
\lambda r(t)\operatorname{mean}\!\left[\sg(gs)h(x,\hat y)\right],
\label{eq:spmr}
\end{equation}
where $r(t)$ is a fixed schedule and $\sg$ denotes stop-gradient. The conflict, target, weight, and strongest competitor are recomputed at every step and detached before the margin is applied. The Forward term therefore remains responsible for observed-label likelihood, while the auxiliary term changes only the selected clean-class comparison. Given $T$, the method uses one network forward pass and $O(BK)$ additional arithmetic for batch size $B$.

\subsection{Role and Stability of the Transition Matrix}
For any supported class $j\neq k$, the reverse posterior favors $j$ over the observed class exactly when
\begin{equation}
q_j>q_k\quad\Longleftrightarrow\quad
z_j-z_k>\log\frac{T_{kk}}{T_{jk}}.
\label{eq:barrier}
\end{equation}
The transition matrix therefore contributes a class-pair barrier in logit space. Under $K$-class symmetric noise with rate $\eta<(K-1)/K$, every posterior conflict selects the model's most probable class, and the barrier reduces to
\begin{equation}
g=\ind[\hat y\neq k]\ind\!\left[
\log\frac{p_{\hat y}}{p_k}>
\log\frac{(1-\eta)(K-1)}{\eta}\right].
\label{eq:symmetric}
\end{equation}
Thus, under symmetric noise, the transition mainly adapts the conflict threshold to the noise rate and class count. Under asymmetric noise, class-specific off-diagonal entries can also change which alternative becomes the target.

The same score representation yields a local stability statement. If the two transition entries involved in $\Delta$ have relative error at most $\rho<1$, then
\begin{equation}
|\widehat\Delta-\Delta|\leq
\log\frac{1+\rho}{1-\rho}.
\label{eq:perturb_bound}
\end{equation}
A pairwise lead larger than this bound keeps its sign under the perturbation. The result does not guarantee an unchanged training trajectory, but it provides a direct scale for interpreting target changes caused by transition estimation; the bound follows by subtracting the two perturbed log-ratios.

\section{Controlled Attribution}
The method contains three coupled choices, so ordinary component removal is not sufficient for attribution. We use matched interventions that change one role at a time and keep the remaining training signal fixed.

\paragraph{Evidence and target.}
For evidence, the fixed-coefficient, prediction-gap, and posterior-gap variants use exactly the same posterior-conflict examples, reverse-posterior targets, pairwise hinge, schedule, and tuning budget. Only the scalar coefficient changes. This comparison asks whether transition-adjusted separation matters beyond ordinary model confidence. For target choice, asymmetric-noise experiments keep the posterior conflict set, coefficient, action, and schedule fixed and replace only $\hat y$ with the model's most probable class. Symmetric noise serves as a useful reference because Eq.~\eqref{eq:symmetric} makes the two targets coincide after a supported conflict.

\paragraph{Update direction.}
Natural auxiliary losses differ in direction, magnitude, saturation, and curvature. We therefore evaluate practical Forward reweighting, hard-target cross-entropy, and the pairwise margin, and then run a separate first-order comparison. For active example $i$, let
\begin{equation}
\begin{aligned}
d_i^{\rm pair}&=e_{c_i^*}-e_{\hat y_i},
&d_i^{\rm hard}&=p_i-e_{\hat y_i},\\
d_i^{\rm F}&=p_i-q_i,
&u_i&=\sg\!\left(g_is_i\ind[h_i>0]\right),\\
\widehat d_i^R&=\sg\!\left(d_i^R/\|d_i^R\|_2\right).
\end{aligned}
\label{eq:directions}
\end{equation}
Directions with norm at most $10^{-6}$ are masked. The detached linear surrogate
\begin{equation}
\widetilde R_{\rm dir}^{R}=\frac{1}{B}\sum_i
u_i\langle\widehat d_i^R,z_i\rangle
\label{eq:dir_surrogate}
\end{equation}
has logit gradient $\frac{u_i}{B}\widehat d_i^R$. The three variants consequently share the active examples, coefficients, targets, and per-example additional logit-gradient norms. Their remaining first-order difference is direction. For a linear classification head $z=W\phi(x)$, $\nabla_W\ell=(\nabla_z\ell)\phi(x)^\top$, so the same comparison also matches the added classification-head gradient norm for each example. It does not equate full-network parameter gradients or long-horizon optimization.

\section{Experiments}
\subsection{Experimental Setup}
The main study uses CIFAR-10 with 50\% and 80\% symmetric noise and 40\% semantic asymmetric noise, together with CIFAR-100 at 50\% and 80\% symmetric noise \cite{krizhevsky2009cifar}. Known-transition experiments isolate the training mechanism. Anchor and DualT provide estimated transition matrices. CIFAR-10N Worst and CIFAR-100N Fine evaluate naturally collected annotations \cite{wei2022cifarn}. For each human-annotation benchmark, one DualT matrix is estimated from the noisy training split and shared by all paired methods. PreActResNet-18 is the primary backbone, and WRN-28-2 is evaluated without retuning \cite{he2016identity,zagoruyko2016wide}.

All main runs use 200 epochs. SPMR is trained for the full schedule without Mixup or early stopping. The five known-transition benchmarks and the central evidence, target, and action comparisons use five paired seeds. Transition-estimation, human-annotation, stronger-recipe, and architecture studies use three paired seeds. Paired methods share the transition artifact, corruption pattern, initialization, data order, and augmentation randomness. Five-seed comparisons report paired differences and bootstrap intervals \cite{efron1979bootstrap}. Three-seed studies report every seed and mean $\pm$ sample standard deviation.

A class-balanced clean set of 1,000 CIFAR-10 images is removed from every method's training split. On a single C10-S60 development setting, it selects one global SPMR coefficient, one Mixup coefficient, one stopping epoch, and one diagnostic prediction threshold. The selected values are transferred unchanged to all reported benchmarks. Nested 250- and 500-image subsets rescore the same checkpoints to measure selection stability. Code, configurations, and run manifests will be released upon publication.

The evaluation is organized around four questions. Table~\ref{tab:main} first asks whether the full method improves Forward and whether the gain survives stronger recipes and transfer settings. Table~\ref{tab:controls} then isolates evidence, target, and action. A transition analysis measures how estimation error propagates through conflict selection, target choice, and weighting. Finally, activity, gradient alignment, and wrong-target update mass test whether the mechanism remains active without systematically amplifying incorrect alternatives.

\subsection{Main Results}
\begin{figure}[t]
\centering
\includegraphics[width=\textwidth]{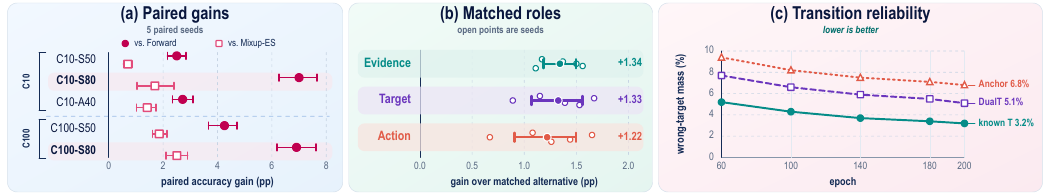}
\caption{\textbf{Summary of performance, attribution, and transition reliability.} (a) Paired gains over full-length Forward (filled circles) and Forward+Mixup-ES (open squares) across known-transition settings. (b) Five-seed gains over the strongest matched alternative for posterior evidence, target choice, and norm-matched action. Open circles are individual seeds, and bars are paired 95\% bootstrap intervals. (c) Cumulative auxiliary-update mass assigned to incorrect targets under known and estimated transitions.}
\label{fig:results}
\end{figure}

Figure~\ref{fig:results} provides a visual summary of the central evidence. Table~\ref{tab:main}(a) reports the five known-transition benchmarks, and Table~\ref{tab:main}(b) summarizes estimated transitions, stronger-recipe compatibility, human annotations, and architecture transfer.

With known transitions, \resulttext{SPMR improves full-length Forward on every benchmark and every paired seed. The gain ranges from 2.50 pp on C10-S50 to 7.00 pp on C10-S80. Relative to Forward+Mixup-ES, the improvement remains positive in all five settings and ranges from 0.70 to 2.51 pp. The largest differences occur at 80\% noise, where full-length Forward exhibits the strongest late-training decline.} Forward and SPMR use the same training images and the full 200-epoch schedule. The early-stopping and Mixup rows are separate training baselines rather than components of SPMR.

\begin{table}[t]
\caption{\textbf{Accuracy under known transitions and across transfer settings.} Panel (a) uses five paired seeds; Figure~\ref{fig:results}(a) shows the paired intervals. Panel (b) reports mean $\pm$ SD for three seeds, with positive gains on all three. Accuracy is in percent and differences are in percentage points (pp).}
\label{tab:main}
\centering
\setlength{\tabcolsep}{2.4pt}
\begin{minipage}[t]{0.61\textwidth}
\vspace{0pt}
\centering\scriptsize
\renewcommand{\arraystretch}{1.18}
\textbf{(a) Known transitions}\par\vspace{2pt}
\begin{tabularx}{\linewidth}{@{}>{\raggedright\arraybackslash}Xccccc@{}}
\toprule
Method & C10-S50 & C10-S80 & C10-A40 & C100-S50 & C100-S80\\
\midrule
Forward & \resultpm{89.1}{0.4} & \resultpm{68.4}{0.8} & \resultpm{87.5}{0.5} & \resultpm{54.7}{0.6} & \resultpm{24.9}{0.7}\\
Forward-ES & \resultpm{90.0}{0.3} & \resultpm{72.0}{0.6} & \resultpm{88.0}{0.4} & \resultpm{56.0}{0.5} & \resultpm{27.8}{0.6}\\
Forward+Mixup-ES & \resultpm{90.9}{0.3} & \resultpm{73.7}{0.6} & \resultpm{88.8}{0.4} & \resultpm{57.1}{0.5} & \resultpm{29.3}{0.6}\\
Noise-Bounded Forward & \resultpm{90.4}{0.3} & \resultpm{71.7}{0.7} & -- & \resultpm{55.8}{0.5} & \resultpm{27.4}{0.6}\\
\textbf{SPMR} & \result{\textbf{91.6$\pm$0.2}} & \result{\textbf{75.4$\pm$0.4}} & \result{\textbf{90.2$\pm$0.2}} & \result{\textbf{59.0$\pm$0.3}} & \result{\textbf{31.8$\pm$0.3}}\\
\midrule
SPMR $-$ Forward (pp) & \result{+2.50} & \result{+7.00} & \result{+2.72} & \result{+4.26} & \result{+6.91}\\
SPMR $-$ Mixup-ES (pp) & \result{+0.70} & \result{+1.70} & \result{+1.41} & \result{+1.86} & \result{+2.51}\\
\bottomrule
\end{tabularx}
\end{minipage}\hfill
\begin{minipage}[t]{0.37\textwidth}
\vspace{0pt}
\centering\scriptsize
\setlength{\tabcolsep}{1.8pt}
\renewcommand{\arraystretch}{0.89}
\textbf{(b) Estimated transitions and transfer}\par\vspace{2pt}
\begin{tabularx}{\linewidth}{@{}>{\raggedright\arraybackslash}Xccc@{}}
\toprule
Setting & Base & +SPMR & $\Delta$ (pp)\\
\midrule
Anchor, C10-S80 & \resultpm{65.8}{0.4} & \resultpm{71.0}{0.4} & \result{+5.20}\\
DualT, C10-S80 & \resultpm{66.7}{0.5} & \resultpm{71.6}{0.4} & \result{+4.85}\\
DualT, C10-A40 & \resultpm{85.5}{0.4} & \resultpm{87.5}{0.3} & \result{+2.01}\\
DualT, C100-S50 & \resultpm{52.0}{0.5} & \resultpm{53.8}{0.5} & \result{+1.78}\\
\midrule
FEC equal-control & \resultpm{86.1}{0.1} & \resultpm{87.2}{0.2} & \result{+1.13}\\
JEC equal-control & \resultpm{65.2}{0.2} & \resultpm{66.2}{0.3} & \result{+1.01}\\
\midrule
C10N Worst & \resultpm{83.5}{0.3} & \resultpm{84.9}{0.3} & \result{+1.43}\\
C100N Fine & \resultpm{58.6}{0.5} & \resultpm{60.0}{0.5} & \result{+1.42}\\
WRN, C10-S80 & \resultpm{67.2}{0.6} & \resultpm{73.5}{0.4} & \result{+6.31}\\
\bottomrule
\end{tabularx}
\end{minipage}
\end{table}

The transfer results address a different question. \resulttext{SPMR remains beneficial with both transition estimators, improves the FEC and JEC equal-control baselines by 1.13 and 1.01 pp, transfers without retuning to WRN-28-2, and improves both human-annotation benchmarks on all three seeds. Full-schedule SPMR+Mixup also outperforms either component.} Given $T$, the training-time overhead is \result{6.4\%}.

\subsection{Posterior Evidence and Transition Effects}
The fixed, prediction-gap, and posterior-gap variants share the conflict set, target, pairwise action, schedule, and coefficient search. \resulttext{The posterior gap is better on all five seeds, increasing target-correctness AUROC from 0.769 to 0.842 and its rank correlation with clean-gradient alignment from 0.329 to 0.486. The leading-pair mass and transition-adjusted separation reach 73.22\% and 74.17\% separately, while their product reaches 75.40\%.} Common confidence alternatives remain weaker under the same intervention.

Under symmetric noise, Eq.~\eqref{eq:symmetric} supplies a threshold that adapts analytically to the noise rate and class count without changing the winning alternative. \resulttext{It transfers better than a scalar threshold selected once on C10-S60 in all four symmetric held-out settings.} Under controlled transition perturbations, \resulttext{off-diagonal errors disrupt the selected target and weighted conflict more than equally sized diagonal errors.}

\subsection{Target and Classifier Action}
Table~\ref{tab:controls} changes one role at a time. Panel (a) contains the matched interventions, while panel (b) compares SPMR with HMW and OGC under the same backbone, augmentation, and training budget.

\begin{table}[H]
\caption{\textbf{Controlled comparisons of posterior evidence, target, and action.} Evidence and action use C10-S80 with five paired seeds; target choice uses C10-A40. Intervals are paired 95\% bootstrap intervals against the strongest matched alternative.}
\label{tab:controls}
\centering
\setlength{\tabcolsep}{3.0pt}
\renewcommand{\arraystretch}{1.08}
\begin{minipage}[t]{0.665\textwidth}
\vspace{0pt}
\centering\scriptsize
\textbf{(a) Matched role interventions}\par\vspace{2pt}
\begin{tabularx}{\linewidth}{@{}L{2.15cm}>{\raggedright\arraybackslash}Xcc@{}}
\toprule
Role & Variant & Accuracy (\%) & SPMR advantage (pp)\\
\midrule
\multirow{3}{*}{Evidence}
 & fixed coefficient & \resultpm{72.6}{0.3} & --\\
 & $p$ top-two gap & \resultpm{74.1}{0.3} & \result{1.34 [1.19, 1.50]}\\
 & \textbf{$q$ top-two gap} & \result{\textbf{75.4$\pm$0.4}} & --\\
\midrule
\multirow{3}{*}{Practical action}
 & Forward reweighting & \resultpm{73.3}{0.3} & --\\
 & hard-target CE & \resultpm{73.8}{0.3} & \result{1.61 [1.14, 2.11]}\\
 & \textbf{pairwise margin} & \result{\textbf{75.4$\pm$0.4}} & --\\
\midrule
\multirow{3}{2.15cm}{Norm-matched action}
 & Forward direction & \resultpm{72.9}{0.3} & --\\
 & hard-target direction & \resultpm{73.6}{0.2} & \result{1.22 [0.92, 1.49]}\\
 & \textbf{pairwise direction} & \result{\textbf{74.8$\pm$0.3}} & --\\
\midrule
\multirow{2}{*}{Target, C10-A40}
 & $q$ conflict, $p$ target & \resultpm{88.9}{0.1} & \result{1.33 [1.07, 1.56]}\\
 & \textbf{$q$ conflict, $q$ target} & \result{\textbf{90.2$\pm$0.2}} & --\\
\bottomrule
\end{tabularx}
\end{minipage}\hfill
\begin{minipage}[t]{0.305\textwidth}
\vspace{0pt}
\centering\scriptsize
\renewcommand{\arraystretch}{1.20}
\textbf{(b) Common-protocol methods}\par\vspace{2pt}
\begin{tabularx}{\linewidth}{@{}>{\raggedright\arraybackslash}Xlc@{}}
\toprule
Setting & Method & Accuracy (\%)\\
\midrule
\multirow{3}{*}{C10-S80}
 & HMW & \resultpm{73.2}{0.5}\\
 & OGC & \resultpm{73.8}{0.4}\\
 & \textbf{SPMR} & \result{\textbf{75.4$\pm$0.4}}\\
\midrule
\multirow{3}{*}{C100-S50}
 & HMW & \resultpm{57.3}{0.4}\\
 & OGC & \resultpm{57.9}{0.4}\\
 & \textbf{SPMR} & \result{\textbf{59.0$\pm$0.3}}\\
\bottomrule
\end{tabularx}
\par\vspace{4pt}
\raggedright\emph{Boldface marks the best result within each setting. Entries are mean $\pm$ SD under the same backbone, augmentation, and training budget.}\par
\end{minipage}
\end{table}
\FloatBarrier
\setlength{\parskip}{0pt}

Under symmetric noise, the transition changes the threshold but not the selected alternative. Under asymmetric noise it can change the target itself. \resulttext{On C10-A40, reverse-posterior and model-only targets differ on a median of 3,180 training examples. The two targets are correct on 74.6\% and 60.8\% of this subset, respectively, and replacing only the target improves accuracy by 1.33 percentage points over five positive seeds.} A cyclic asymmetric replication gives the same target ordering.

Practical losses show that the pairwise margin is more effective than conflict-weighted Forward or dense hard-target cross-entropy at their natural scales. The direction study then removes the per-example norm difference. \resulttext{After matching active examples, coefficients, targets, and logit-gradient norms, the pairwise direction remains 1.22 percentage points above the hard-target direction and 1.91 percentage points above the Forward direction. The ordering persists in the classification head, last residual block, and full-network clean-gradient diagnostics at epochs 100 and 200.} Alignment decreases with depth, so the logit result is a local explanation rather than a complete model of deep-network optimization.

\par\noindent\textbf{Reliability and efficiency.} After the 40-epoch warm-up and 20-epoch ramp, \resulttext{conflict coverage stabilizes, update mass peaks near epoch 100 and then decreases, and invalid directions remain below 0.2\%.} Without retuning, the WRN-28-2 audit preserves similar conflict and margin activity despite a larger logit norm. Transition estimation is the clearer reliability boundary. \resulttext{Cumulative wrong-target update mass is 3.2\% with known $T$, 5.1\% with DualT, and 6.8\% with Anchor, and each curve decreases late in training.} Figure~\ref{fig:results}(c) reports the temporal wrong-target update-mass diagnostic. Given $T$, SPMR reuses the Forward pass, adds $O(BK)$ arithmetic and $O(K^2)$ transition storage, and increases training time by \result{6.4\%}.

\section{Discussion}
The results distinguish two roles of the transition model. Under symmetric noise it mainly provides an analytic, setting-dependent conflict threshold. Under asymmetric noise it can also reorder candidate classes, so estimation error propagates through evidence, target, and update strength. The perturbation analysis applies while the relevant transition entries share support. Support-changing errors remain a harder case and motivate the explicit target-disagreement and wrong-update diagnostics.

These distinctions also clarify comparison with broader noisy-label systems. DivideMix and UniCon jointly change the data partition, representation learning, and objective, so they are system-level alternatives rather than matched controls for a local Forward intervention. The FEC/JEC and full-schedule Mixup studies instead test complementarity after the surrounding recipe is strengthened. With estimated $T$, target disagreement across estimators provides a useful warning signal.

SPMR is a finite-sample regularizer around Forward correction rather than a new population-risk guarantee. Proposition~1 describes the local action of an active margin. More formally, a Forward stationary point remains stationary for the detached combined field whenever the selected margins are already satisfied, but active margins can change other stationary points. \resulttext{Accuracy also remains within 0.8 percentage points across margins from 0.5 to 2 and warm-ups from 20 to 60 epochs, with clearer degradation only at extreme coefficients.} Scaling this analysis to larger visual corpora and heterogeneous robotic data is an important next step \cite{zitkovich2023rt2,openx2024}.

\section{Conclusion}
SPMR preserves the Forward likelihood while translating reverse-posterior disagreement into a selective pairwise update on the clean logits. The evidence--target--action decomposition separates posterior information from the intervention it triggers, and \resulttext{matched comparisons support each role across the evaluated transitions and training settings.} The approach uses latent-class evidence without turning every disagreement into a replacement label.

\begingroup
\small
\setlength{\bibsep}{2pt plus 0.3ex}
\bibliographystyle{plainnat}
\bibliography{references}

@inproceedings{patrini2017making,
  title={Making Deep Neural Networks Robust to Label Noise: A Loss Correction Approach},
  author={Patrini, Giorgio and Rozza, Alessandro and Menon, Aditya Krishna and Nock, Richard and Qu, Lizhen},
  booktitle={Proceedings of the IEEE Conference on Computer Vision and Pattern Recognition},
  pages={1944--1952},
  year={2017}
}

@inproceedings{feng2026deconstructing,
  title={Deconstructing the Failure of Ideal Noise Correction: A Three-Pillar Diagnosis},
  author={Feng, Chen and Zhi, Zhuo and Huang, Zhao and Ge, Jiawei and Xiao, Ling and Sebe, Nicu and Tzimiropoulos, Georgios and Patras, Ioannis},
  booktitle={Proceedings of the IEEE/CVF Conference on Computer Vision and Pattern Recognition},
  pages={34512--34523},
  year={2026}
}

@article{toner2023forward,
  title={Label Noise: Correcting the Forward-Correction},
  author={Toner, William and Storkey, Amos},
  journal={arXiv preprint arXiv:2307.13100},
  year={2023}
}

@inproceedings{guo2024cdro,
  title={Learning from Noisy Labels via Conditional Distributionally Robust Optimization},
  author={Guo, Hui and Yi, Grace Y. and Wang, Boyu},
  booktitle={Advances in Neural Information Processing Systems},
  volume={37},
  year={2024}
}

@article{zhang2024hmw,
  title={Learning with Noisy Labels Using Hyperspherical Margin Weighting},
  author={Zhang, Shuo and Li, Yuwen and Wang, Zhongyu and Li, Jianqing and Liu, Chengyu},
  journal={Proceedings of the AAAI Conference on Artificial Intelligence},
  volume={38},
  number={15},
  pages={16848--16856},
  year={2024}
}

@article{ye2025ogc,
  title={Optimized Gradient Clipping for Noisy Label Learning},
  author={Ye, Xichen and Wu, Yifan and Zhang, Weizhong and Li, Xiaoqiang and Chen, Yifan and Jin, Cheng},
  journal={Proceedings of the AAAI Conference on Artificial Intelligence},
  volume={39},
  number={9},
  pages={9463--9471},
  year={2025}
}

@inproceedings{yao2020dualt,
  title={Dual T: Reducing Estimation Error for Transition Matrix in Label-noise Learning},
  author={Yao, Yu and Liu, Tongliang and Han, Bo and Gong, Mingming and Deng, Jiankang and Niu, Gang and Sugiyama, Masashi},
  booktitle={Advances in Neural Information Processing Systems},
  volume={33},
  pages={7260--7271},
  year={2020}
}

@inproceedings{xia2019revision,
  title={Are Anchor Points Really Indispensable in Label-Noise Learning?},
  author={Xia, Xiaobo and Liu, Tongliang and Wang, Nannan and Han, Bo and Gong, Chen and Niu, Gang and Sugiyama, Masashi},
  booktitle={Advances in Neural Information Processing Systems},
  volume={32},
  year={2019}
}

@inproceedings{liu2020elr,
  title={Early-Learning Regularization Prevents Memorization of Noisy Labels},
  author={Liu, Sheng and Niles-Weed, Jonathan and Razavian, Narges and Fernandez-Granda, Carlos},
  booktitle={Advances in Neural Information Processing Systems},
  volume={33},
  pages={20331--20342},
  year={2020}
}

@inproceedings{li2020dividemix,
  title={DivideMix: Learning with Noisy Labels as Semi-supervised Learning},
  author={Li, Junnan and Socher, Richard and Hoi, Steven C. H.},
  booktitle={International Conference on Learning Representations},
  year={2020}
}

@inproceedings{han2018coteaching,
  title={Co-teaching: Robust Training of Deep Neural Networks with Extremely Noisy Labels},
  author={Han, Bo and Yao, Quanming and Yu, Xingrui and Niu, Gang and Xu, Miao and Hu, Weihua and Tsang, Ivor and Sugiyama, Masashi},
  booktitle={Advances in Neural Information Processing Systems},
  volume={31},
  year={2018}
}

@inproceedings{zhang2018mixup,
  title={mixup: Beyond Empirical Risk Minimization},
  author={Zhang, Hongyi and Cisse, Moustapha and Dauphin, Yann N. and Lopez-Paz, David},
  booktitle={International Conference on Learning Representations},
  year={2018}
}

@inproceedings{wei2022cifarn,
  title={Learning with Noisy Labels Revisited: A Study Using Real-World Human Annotations},
  author={Wei, Jiaheng and Zhu, Zhaowei and Cheng, Hao and Liu, Tongliang and Niu, Gang and Liu, Yang},
  booktitle={International Conference on Learning Representations},
  year={2022}
}

@inproceedings{he2016identity,
  title={Identity Mappings in Deep Residual Networks},
  author={He, Kaiming and Zhang, Xiangyu and Ren, Shaoqing and Sun, Jian},
  booktitle={European Conference on Computer Vision},
  pages={630--645},
  year={2016}
}

@inproceedings{zagoruyko2016wide,
  title={Wide Residual Networks},
  author={Zagoruyko, Sergey and Komodakis, Nikos},
  booktitle={British Machine Vision Conference},
  year={2016}
}

@inproceedings{natarajan2013learning,
  title={Learning with Noisy Labels},
  author={Natarajan, Nagarajan and Dhillon, Inderjit S. and Ravikumar, Pradeep and Tewari, Ambuj},
  booktitle={Advances in Neural Information Processing Systems},
  volume={26},
  year={2013}
}

@inproceedings{arazo2019pencil,
  title={Unsupervised Label Noise Modeling and Loss Correction},
  author={Arazo, Eric and Ortego, Diego and Albert, Paul and O'Connor, Noel E. and McGuinness, Kevin},
  booktitle={International Conference on Machine Learning},
  pages={312--321},
  year={2019}
}

@inproceedings{ma2020normalized,
  title={Normalized Loss Functions for Deep Learning with Noisy Labels},
  author={Ma, Xingjun and Huang, Hanxun and Wang, Yisen and Romano, Simone and Erfani, Sarah and Bailey, James},
  booktitle={International Conference on Machine Learning},
  pages={6543--6553},
  year={2020}
}

@inproceedings{englesson2021gjs,
  title={Generalized Jensen-Shannon Divergence Loss for Learning with Noisy Labels},
  author={Englesson, Erik and Azizpour, Hossein},
  booktitle={Advances in Neural Information Processing Systems},
  volume={34},
  pages={30284--30297},
  year={2021}
}

@inproceedings{reed2015bootstrapping,
  title={Training Deep Neural Networks on Noisy Labels with Bootstrapping},
  author={Reed, Scott and Lee, Honglak and Anguelov, Dragomir and Szegedy, Christian and Erhan, Dumitru and Rabinovich, Andrew},
  booktitle={International Conference on Learning Representations Workshop},
  year={2015}
}

@inproceedings{gosh2017robust,
  title={Robust Loss Functions under Label Noise for Deep Neural Networks},
  author={Ghosh, Aritra and Kumar, Himanshu and Sastry, P. S.},
  booktitle={Proceedings of the AAAI Conference on Artificial Intelligence},
  volume={31},
  year={2017}
}

@techreport{krizhevsky2009cifar,
  title={Learning Multiple Layers of Features from Tiny Images},
  author={Krizhevsky, Alex},
  institution={University of Toronto},
  year={2009}
}

@article{efron1979bootstrap,
  title={Bootstrap Methods: Another Look at the Jackknife},
  author={Efron, Bradley},
  journal={The Annals of Statistics},
  volume={7},
  number={1},
  pages={1--26},
  year={1979}
}

@inproceedings{sukhbaatar2015training,
  title={Training Convolutional Networks with Noisy Labels},
  author={Sukhbaatar, Sainbayar and Bruna, Joan and Paluri, Manohar and Bourdev, Lubomir and Fergus, Rob},
  booktitle={International Conference on Learning Representations Workshop},
  year={2015}
}

@inproceedings{goldberger2017noise,
  title={Training Deep Neural-Networks Using a Noise Adaptation Layer},
  author={Goldberger, Jacob and Ben-Reuven, Ehud},
  booktitle={International Conference on Learning Representations},
  year={2017}
}

@inproceedings{hendrycks2018trusted,
  title={Using Trusted Data to Train Deep Networks on Labels Corrupted by Severe Noise},
  author={Hendrycks, Dan and Mazeika, Mantas and Wilson, Duncan and Gimpel, Kevin},
  booktitle={Advances in Neural Information Processing Systems},
  volume={31},
  year={2018}
}

@inproceedings{yu2019coteachingplus,
  title={How Does Disagreement Help Generalization against Label Corruption?},
  author={Yu, Xingrui and Han, Bo and Yao, Jiangchao and Niu, Gang and Tsang, Ivor W. and Sugiyama, Masashi},
  booktitle={International Conference on Machine Learning},
  volume={97},
  pages={7164--7173},
  year={2019}
}

@inproceedings{wei2020jocor,
  title={Combating Noisy Labels by Agreement: A Joint Training Method with Co-Regularization},
  author={Wei, Hongxin and Feng, Lei and Chen, Xiangyu and An, Bo},
  booktitle={Proceedings of the IEEE/CVF Conference on Computer Vision and Pattern Recognition},
  pages={13726--13735},
  year={2020}
}

@inproceedings{nguyen2020self,
  title={SELF: Learning to Filter Noisy Labels with Self-Ensembling},
  author={Nguyen, Duc Tam and Mummadi, Chaithanya Kumar and Ngo, Thi Phuong Nhung and Nguyen, Thi Hoai Phuong and Beggel, Laura and Brox, Thomas},
  booktitle={International Conference on Learning Representations},
  year={2020}
}

@inproceedings{pleiss2020aum,
  title={Identifying Mislabeled Data Using the Area Under the Margin Ranking},
  author={Pleiss, Geoff and Zhang, Tianyi and Elenberg, Ethan R. and Weinberger, Kilian Q.},
  booktitle={Advances in Neural Information Processing Systems},
  volume={33},
  pages={17044--17056},
  year={2020}
}

@article{northcutt2021confident,
  title={Confident Learning: Estimating Uncertainty in Dataset Labels},
  author={Northcutt, Curtis G. and Jiang, Lu and Chuang, Isaac L.},
  journal={Journal of Artificial Intelligence Research},
  volume={70},
  pages={1373--1411},
  year={2021}
}

@inproceedings{zhang2018gce,
  title={Generalized Cross Entropy Loss for Training Deep Neural Networks with Noisy Labels},
  author={Zhang, Zhilu and Sabuncu, Mert R.},
  booktitle={Advances in Neural Information Processing Systems},
  volume={31},
  year={2018}
}

@inproceedings{wang2019sce,
  title={Symmetric Cross Entropy for Robust Learning with Noisy Labels},
  author={Wang, Yisen and Ma, Xingjun and Chen, Zaiyi and Luo, Yuan and Yi, Jinfeng and Bailey, James},
  booktitle={Proceedings of the IEEE/CVF International Conference on Computer Vision},
  pages={322--330},
  year={2019}
}

@inproceedings{xia2021robustearly,
  title={Robust Early-Learning: Hindering the Memorization of Noisy Labels},
  author={Xia, Xiaobo and Liu, Tongliang and Han, Bo and Gong, Chen and Wang, Nannan and Ge, Zongyuan and Chang, Yi},
  booktitle={International Conference on Learning Representations},
  year={2021}
}

@inproceedings{nishi2021augmentation,
  title={Augmentation Strategies for Learning with Noisy Labels},
  author={Nishi, Kento and Ding, Yi and Rich, Alex and Hollerer, Tobias},
  booktitle={Proceedings of the IEEE/CVF Conference on Computer Vision and Pattern Recognition},
  pages={8022--8031},
  year={2021}
}

@inproceedings{kim2023crosssplit,
  title={CrossSplit: Mitigating Label Noise Memorization through Data Splitting},
  author={Kim, Jihye and Baratin, Aristide and Zhang, Yan and Lacoste-Julien, Simon},
  booktitle={International Conference on Machine Learning},
  pages={16377--16392},
  year={2023}
}

@inproceedings{xue2022contrastive,
  title={Investigating Why Contrastive Learning Benefits Robustness against Label Noise},
  author={Xue, Yihao and Whitecross, Kyle and Mirzasoleiman, Baharan},
  booktitle={International Conference on Machine Learning},
  pages={24851--24871},
  year={2022}
}

@article{song2023survey,
  author  = {Song, Hwanjun and Kim, Minseok and Park, Dongmin and Shin, Yooju and Lee, Jae-Gil},
  title   = {Learning from Noisy Labels with Deep Neural Networks: A Survey},
  journal = {IEEE Transactions on Neural Networks and Learning Systems},
  volume  = {34},
  number  = {11},
  pages   = {8135--8153},
  year    = {2023},
  doi     = {10.1109/TNNLS.2022.3152527}
}

@inproceedings{karim2022unicon,
  author    = {Karim, Nazmul and Rizve, Mamshad Nayeem and Rahnavard, Nazanin and Mian, Ajmal and Shah, Mubarak},
  title     = {{UniCon}: Combating Label Noise Through Uniform Selection and Contrastive Learning},
  booktitle = {Proceedings of the IEEE/CVF Conference on Computer Vision and Pattern Recognition},
  pages     = {9676--9686},
  year      = {2022}
}

@misc{yao2022posteriorselection,
  author       = {Yao, Yu and Li, Xuefeng and Liu, Tongliang and Blair, Alan and Gong, Mingming and Han, Bo and Niu, Gang and Sugiyama, Masashi},
  title        = {Can Label-Noise Transition Matrix Help to Improve Sample Selection and Label Correction?},
  year         = {2022},
  howpublished = {Withdrawn ICLR submission, OpenReview},
  note         = {OpenReview identifier c0AD3ll9Wyv}
}

@inproceedings{zitkovich2023rt2,
  author    = {Zitkovich, Brianna and others},
  title     = {{RT-2}: Vision-Language-Action Models Transfer Web Knowledge to Robotic Control},
  booktitle = {Proceedings of the 7th Conference on Robot Learning},
  series    = {Proceedings of Machine Learning Research},
  volume    = {229},
  pages     = {2165--2183},
  year      = {2023}
}

@inproceedings{openx2024,
  author    = {{Open X-Embodiment Collaboration}},
  title     = {Open {X-Embodiment}: Robotic Learning Datasets and {RT-X} Models},
  booktitle = {2024 IEEE International Conference on Robotics and Automation},
  pages     = {6892--6903},
  year      = {2024},
  doi       = {10.1109/ICRA57147.2024.10611477}
}
\endgroup
\end{document}